 \documentclass[final,3p,times,twocolumn]{elsarticle}

\makeatletter \def\ps@pprintTitle{  \let\@oddhead\@empty  \let\@evenhead\@empty  \def\@oddfoot{\hfill\thepage}  \def\@evenfoot{\thepage\hfill}} \makeatother

\usepackage{amssymb}
\usepackage{amsthm}
\usepackage{amsmath}
\usepackage{hyperref}
\usepackage{float}
\usepackage{proof}
\usepackage{listings}
\usepackage{xcolor}
\usepackage{multirow}

\definecolor{codegreen}{rgb}{0,0.6,0}
\definecolor{codegray}{rgb}{0.5,0.5,0.5}
\definecolor{codepurple}{rgb}{0.58,0,0.82}
\definecolor{backcolour}{rgb}{0.95,0.95,0.92}

\lstdefinestyle{mystyle}{
    backgroundcolor=\color{backcolour},   
    commentstyle=\color{codegreen},
    keywordstyle=\color{magenta},
    numberstyle=\tiny\color{codegray},
    stringstyle=\color{codepurple},
    basicstyle=\ttfamily\footnotesize,
    breakatwhitespace=false,         
    breaklines=true,                 
    captionpos=b,                    
    keepspaces=true,                 
    numbers=left,                    
    numbersep=5pt,                  
    showspaces=false,                
    showstringspaces=false,
    showtabs=false,                  
    tabsize=2
}

\journal{International Journal of Approximate Reasoning}

\begin{document}

\begin{frontmatter}



\title{Finding, Scoring and Explaining Arguments in Bayesian Networks}


\author[University of Aberdeen]{Jaime Sevilla}

\affiliation[University of Aberdeen]{organization={Department of Computer Science, Aberdeen University},
            addressline={King's College}, 
            city={Aberdeen},
            postcode={AB24 3FX}, 
            state={Scotland},
            country={United Kingdom}}

\begin{abstract}
We propose a new approach to explain Bayesian Networks. The approach revolves around a new definition of a probabilistic argument and the evidence it provides. We define a notion of  independent arguments, and propose an algorithm to extract a list of relevant, independent arguments given a Bayesian Network, a target node and a set of observations. To demonstrate the relevance of the arguments, we show how we can use the extracted arguments to approximate message passing. Finally, we show a simple scheme to explain the arguments in natural language.
\end{abstract}




\end{frontmatter}


\section{Introduction}
\label{sec:introduction}
Explaining statistical reasoning is a difficult task. 

Bayesian Models have nodes with clear interpretations and parameters with a clear interpretation in term of conditional probabilities. 

And yet large networks are daunting. There is no obvious way of finding the relevant parts of the network to explain to the user.

In this article we will explore a way of relating the available evidence to the outcome of a Bayesian Network - what we will call \textbf{arguments}. First we will explain \textbf{what an argument is} in our framework and \textbf{how to evaluate the conclusions of an argument} using a novel definition of the \textbf{effects of an inference step}. We will use this to delineate \textbf{when it is appropriate to break a complex argument} into simpler, more modular arguments. Armed with these two conceptual tools we will explain \textbf{how to find the most relevant arguments} that apply given some evidence and a target outcome of interest. To show that our argument selection is relevant, we will also check \textbf{how well can we approximate message passing} using our argument finding and argument scoring tools. And finally we will show \textbf{how to explain an argument} using natural language. 

An interactive version of this article can be found in the associated open source package
\href{https://gitlab.nl4xai.eu/jaime.sevilla/explainbn}{ExplainBN}\footnote{https://gitlab.nl4xai.eu/jaime.sevilla/explainbn}. This package also contains an implementation of all the algorithms discussed through this article. The packages is implemented in Python, and relies on the PGMPy open source software for manipulating Bayesian Networks \cite{ankan_pgmpy_2015}.

\begin{figure}
    \centering
    \includegraphics[width=\linewidth]{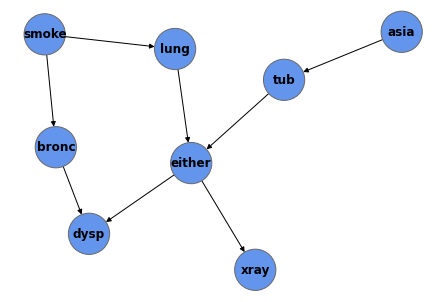}
    \caption{An example of a Bayesian Network, the ASIA Network \cite{lauritzen_local_1988}.}
    \label{fig:asia_network}
\end{figure}

It will be useful to give an example of what we are building towards. Suppose we are working with the Bayesian Network in figure \ref{fig:asia_network}. We are given the observations $\text{xray} = \text{yes}$, $\text{tub} = \text{no}$ and $\text{bronc} = \text{no}$. And we want to figure out the value of node $\text{lung}$.

Then our framework can be used to identify which are the main ways that these observations relate to the target node $\text{lung}$.

In this concrete example the algorithm we have developed would identify two relevant arguments. The explanation produced would be as follows:

\noindent\fbox{\parbox{\linewidth}{
We have observed that the lung xray shows an abnormality and the patient does not have tuberculosis.\\
That the lung xray shows an abnormality is evidence that the patient has a lung disease (strong inference).\\
That the patient has a lung disease and the patient does not have tuberculosis is evidence that the patient has lung cancer (strong inference).\\

We have observed that the patient has bronchitis.\\
That the patient has bronchitis is evidence that the patient smokes (moderate inference).\\
That the patient smokes causes that the patient has lung cancer (weak inference).
}}

\section{Preliminaries}

\textbf{Bayesian Networks} and the associated algorithms are a way of efficiently representing and performing probabilistic inference on joint probability distributions over many variables.

A Bayesian Network is composed of a collection of conditional probability tables $P(Y=y | X_1=x_1, ..., X_M=x_M)$, such that the relation $\text{Parents}(Y) = \{X_1, ..., X_n\}$ defines a directed acyclic graph between the variables.

The joint distribution represented by the Bayesian Network is the product of these conditional probability tables $P(X_1=x_1, ..., X_N = x_n) = \prod_i P(X_i=x_i | \text{Parents}(X_i) = x)$.

We will represent conditional probability tables, observations, distributions over variables and evidence over variable as \textbf{factors}. Formally, a factor is a function from the possible valuations of one or more random variables (its \textit{scope}) to positive real numbers.

\begin{figure}
    \centering
    \includegraphics[width=0.3\linewidth]{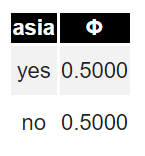}
    \includegraphics[width=0.3\linewidth]{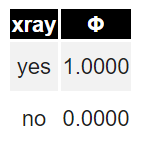}
    \includegraphics[width=0.3\linewidth]{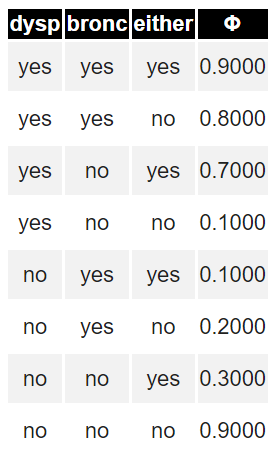}
    \caption{From left to right: a constant factor representing a uniform distribution over the node $\text{asia}$, a lopsided factor representing the observation $\text{xray} = \text{yes}$ and a factor representing the conditional probability table of $\text{either}$ given $\text{dysp}$ and $\text{either}$.}
    \label{fig:factors}
\end{figure}

Factors can be multiplied, divided, marginalized and normalized. See \cite{koller_probabilistic_2009} for more information on factor operations.

The \textbf{factor graph} associated to a Bayesian Network is a bipartite graph where in one side we have all variables of the model and in the other the conditional probability tables (the factors). Each variable is linked to the factors it participates in.

\begin{figure}
    \centering
    \includegraphics[width=\linewidth]{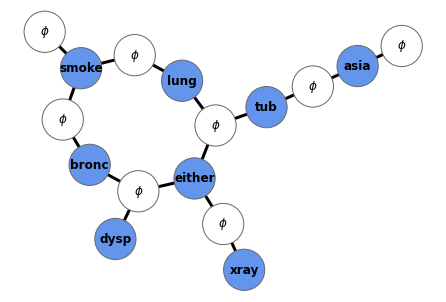}
    \caption{Factor graph associated with the ASIA network. Each of the white nodes is a factor representing a conditional probability table whose scope contains the adjacent nodes in the factor graph.}
    \label{fig:factor_graph}
\end{figure}

Our framework is based on the \textbf{message passing} algorithm for inference in a Bayesian Network. The goal of this algorithm is to compute queries of the form $P(T=t | E_1=e_1,..., E_n=e_n)$. That is, given some evidence $E_1=e_1, ..., E_n = e_n$ we want to compute the posterior beliefs about some variable $T$ in the network.

To perform message passing we start by defining an *initial potential* $\phi_X$ associated with each node $X$ in the corresponding factor graph. For the factor nodes this is easy - we just use the respective conditional probability table. For variable nodes we define either a constant factor or a lopsided factor if the variable was observed. 

Then, we are going to initialize two messages $\delta_{A,B}, \delta_{B,A}$ for each edge $A-B$ in the factor graph. These messages will be used to communicate information across the network.

Each message connects one variable to one factor or viceversa. Formally, messages are factors whose scope will be the scope of the variable - which corresponds to the intersection of the scopes of the nodes it connects. 

At the beginning, each message will be initialized as a constant factor. But this will quickly change as we iteratively apply the message passing equations to compute their new values:  

$$
\delta_{X,Y} \leftarrow \sum_{\text{Scope}(X) \setminus \text{Scope}(Y)} \phi_X \prod_{Z\not = Y} \delta_{Z,X}
$$

In words, we:
\begin{itemize}
    \item Take the factor $\phi_X$ associated with the \textit{sender} node $X$.
    \item We multiply it by all messages being sent to $X$, \textit{except} the message coming from the "receiver" $Y$.
    \item Finally we marginalize all nodes not in the intersection of scopes of the sender and receiver nodes.
\end{itemize}

The update rule is applied until each message reaches an (approximate) fixed point. At that point, the marginal distribution of variable $X$ given the observations can be retrieved from the product of all messages going into $X$, ie $\beta(X) = \prod_Z \delta_{Z,X}$.

Message passing is an exact inference algorithm for polytrees (Bayesian networks with no loops), and an approximate one in general. The order of the updates can affect performance, and several scheduling procedures have been designed. An in-depth exploration of message passing can be found in \cite{koller_probabilistic_2009} or \cite{mooij_understanding_2004}.

\section{Previous work}

There has been plenty of work on explaining Bayesian Networks. See \cite{lacave_review_2002} or \cite{hennessy_explaining_2020} for a review of explanation methods in Bayesian Networks.

Our approach builds on previous work on extracting \textit{chains of reasoning} from Bayesian Networks. \cite{suermondt_explanation_1992} described a method to extract such arguments in their INSITE system. The method was later refined in \cite{haddawy_banter_1997} and \cite{kyrimi_incremental_2020}. Their approaches suggest how to measure and explain the effect of the available evidence on a target node, but it is quite limited when it comes to explain  interactions between chains of reasoning.

We also expand previous work to explain Bayesian Networks using argumentation theory. \cite{vreeswijk_argumentation_2005} is one such paper on the topic, where each conditional probability table in a Bayesian Network is translated as a rule. The resulting argumentation scheme performs an operation similar to Maximum A posterior Probability inference.

\cite{keppens_argument_2012} proposes another approach, where an initial argumentation diagram is refined and labelled to capture some of the interactions occurring in a Bayesian Network. His discussion of convergent and linked arguments led us to propose an alternative characterization of argument independence.

More recently, \cite{timmer_two-phase_2017} described another approach for extracting an argument diagram from a Bayesian Network using what they call \textit{support graphs}, that relate all available evidence to the target in a series of inference steps of the form $X_1=x_1, ..., X_N = x_N \implies Y=y$. Their approach is limited in that it doesn't disentangle the separate effect of the different premises $X_1=x_1, ..., X_N = x_N$ of each step on the rule conclusion $Y=y$ - we address this problem.

To measure the strength of each argument and inference step we favor a notion of logarithmic odds, reminiscent of \cite{madigan_graphical_1997}'s proposal to use the likelihood ratio of the hypothesis given the evidence to quantify the strength of a chain of reasoning. We however develop a notion of strength directly based on the conditional probability tables of the Bayesian Network, which allows us to talk about the strength of specific statistical relations rather than the strength of (a subset of) the evidence variables.

This difference - that we focus on explaining which relations between variables are more important rather than explaining which variables are more important overall - is also what distinguishes our approach to modern model-agnostic explainability frameworks like LIME \cite{ribeiro_why_2016} and SHAP \cite{lundberg_unified_2017}. For example, while SHAP can be useful to detect that bronchitis is important to determine if the patient has dyspnea in figure \ref{fig:asia_network}, it does not offer a direct answer to the question of whether the path $\text{Bronc} \rightarrow \text{Dysp}$ is more important than the path $\text{Bronc} \leftarrow \text{Smoking} \rightarrow \text{Lung} \rightarrow \text{Either} \rightarrow \text{Dysp}$ to explain the outcome.  

\section{Arguments in a Bayesian Network}

Our goal in this tutorial is to take a Bayesian network and produce a list of considerations that relate the observations we have made to some unobserved variables we are interested in. These considerations we will call \textbf{arguments}.

An informal example of an argument might be:

\noindent\fbox{\parbox{\linewidth}{
We have observed that the patient does not have tuberculosis and the lung xray shows an abnormality. \\
That the lung xray shows an abnormality is evidence that the patient has a lung disease. \\
That the patient does not have tuberculosis and the patient has a lung disease is evidence that the patient has lung cancer.
}}

We can already see two crucial elements of an argument: the observations or \textbf{premises} and some \textbf{rules} that tell us, given the premises, how to change our beliefs about other variables.

More formally, a \textbf{factor argument} in our framework is going to be a combination of \textit{premises} and of \textit{rules}. Each of the premises represents some \textit{evidence} about a variable, while each \textit{rule} is a function that takes as input evidence about some variables and produces evidence about another variable. Both premises and rules will be represented as \textit{factors} -  lopsided factors over single variables for the former and conditional probability tables for the latter. We call them \textit{factor arguments} to distinguish them from arguments in classical argumentation theory; since we do not deal with classical arguments in this paper we will refer to \textit{factor arguments} simply as \textbf{arguments}.

\textbf{EXAMPLE ARGUMENT:}

$$
\begin{array}{cc}
\text{Premises}&\text{Rules}\\
\Delta_{Xray}&\Delta_{Xray} \implies_{\phi_1} \Delta_{Either} \\
\Delta_{Tub}&\Delta_{Either}, \Delta_{Tub}  \implies_{\phi_2} \Delta_{Lung}
\end{array}
$$

For our purposes, a more convenient computational representation of arguments will be as a \textbf{directed acyclic graph over a factor graph}. The sources of the graph will be the premises, while each factor in the graph will be one of the rules. The factor nodes will have incoming arrows emanating from the variables involved in the premises of the associated rule, and an outwards arrow pointing towards the variable relevant for the conclusion of the rule.

\begin{figure}[H]
    \centering
    \includegraphics[width=\linewidth]{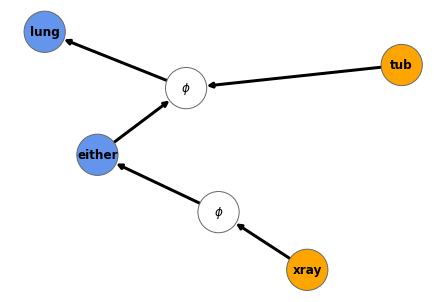}
    \caption{Example of an argument in the ASIA network. The observations of the argument are $\text{Xray} = \text{Yes}$ and $\text{Tub} = \text{No}$. The target is $\text{Lung}$.}
    \label{fig:argument_example}
\end{figure}

Figure \ref{fig:argument_example} is an example of an argument. A description in natural language of the argument might be:

\noindent\fbox{\parbox{\linewidth}{
We have observed that the patient does not have tuberculosis and the lung xray shows an abnormality.  \\
That the lung xray shows an abnormality is evidence that the patient has a lung disease.  \\
That the patient does not have tuberculosis and the patient has a lung disease is evidence that the patient has lung cancer.  
}}

\begin{figure}[H]
    \centering
    \includegraphics[width=\linewidth]{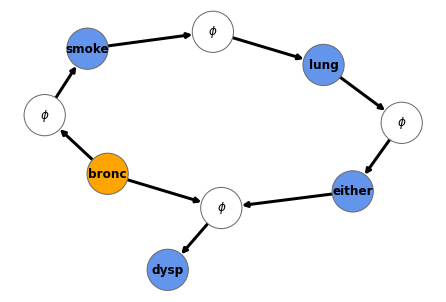}
    \caption{Example of an argument containing a loop in the ASIA network. The premise is that $\text{Bronc} = \text{Yes}$. The target is $\text{Dysp}$.}
    \label{fig:loopy_argument}
\end{figure}

Arguments have to be acyclic, but their skeleton can contain loops. See figure \ref{fig:loopy_argument} for such an example. A description in words of this argument might be

\noindent\fbox{\parbox{\linewidth}{
We have observed that the patient has bronchitis.\\
That the patient has bronchitis is evidence that the patient smokes.\\
That the patient smokes causes that the patient has lung cancer.  \\
That the patient has lung cancer causes that the patient has a lung disease. \\
That the patient has bronchitis and the patient has a lung disease causes that the patient is experiencing shortness of breath.
}}

In an sense, what we will be doing is identifying the most relevant subgraphs in the factor graph one would need to take into account to explain what the message passing algorithm is doing to compute the output.

\section{Effects of an argument}

We would like to calculate how an argument should affect our beliefs about the variables involved in it. We will work towards a definition of this: the \textbf{effects of an argument}.

\subsection{Effect of an inference step}

The center piece of our framework is the definition of the effect of each inference step in an argument. That is, given a factor rule, some factor evidence and a target node, how do we compute the factor evidence of the target node implied by the factor rule and the premises?

Remember than a rule will be represented by a factor $\phi$ - a conditional probability table over some variables. The premises will be factors $\{\Delta_{1}, ..., \Delta_{N}\}$ - representing the evidence over the input variables. And the goal will be to define how this evidence will affect our beliefs about a variable $Y$.

$$
\infer{\Delta_Y}{\Delta_1, ..., \Delta_N & \phi}
$$

We need to specify how exactly will we combine the premise factors and the rule factors to produce the conclusion. After some experimentation we have opted to define this as:

\begin{equation*}
\begin{split}
    \Delta_Y = \text{StepEffect}(\phi, \{\Delta_{1}, ..., \Delta_{N}\}, Y) := \\
\frac{\sum_{Z\not = Y} \phi \cdot \Delta_1 \cdot ... \cdot \Delta_n}{\sum_{Z\not = Y} \phi}
\end{split}
\end{equation*}

The definition is similar to the message passing equations, though we divide by the marginalized factor to separate the information derived from the updates $\Delta_1, ..., \Delta_N$ and the information passively contained in the conditional probability table $\phi$.

\subsection{Effects of an argument}

We would like to be able to discuss the \textbf{effect of an argument} on the variables involved in it. 

For that we will recursively apply the definition of $\text{StepEffect}$ to elucidate the effect of an argument from the evidence up to the target. If a node appears in the conclusion of multiple inference steps, we multiplying the effect of all the steps to arrive at the final effect of the argument on the node.

That is, we define the effect of the argument on each evidence node $E=e_o$ as a lopsided factor that favors said observation.

And we define the effect of the argument on the rest of the variables as:

\begin{equation*}
\begin{split}
\text{ArgumentEffect}(\mathcal{A}, Y) := \Delta_Y^{\mathcal{A}} = \\
\prod_{\phi \in \text{Pred}_\mathcal{A}(Y)} \text{StepEffect}(\phi, \{\Delta_X^{\mathcal{A}} : X \in \text{Pred}_{\mathcal{A}}(\phi)\}, Y) = \\
\prod_{\phi \in \text{Pred}_\mathcal{A}(Y)} 
\frac{\sum_{Z\not = Y} \phi \cdot \prod_{X \in \text{Pred}_{\mathcal{A}}(\phi)}\Delta_X^{\mathcal{A}}}{\sum_{Z\not = Y} \phi}
\end{split}
\end{equation*}

where $\text{Pred}_{\mathcal{A}}(\phi)$ are the parents of the factor $\phi$ in the argument $\mathcal{A}$.

\begin{figure}[H]
    \centering
    \includegraphics[width=0.2\linewidth]{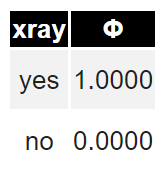}
    \includegraphics[width=0.2\linewidth]{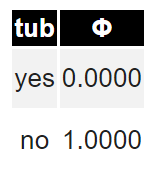}
    \includegraphics[width=0.2\linewidth]{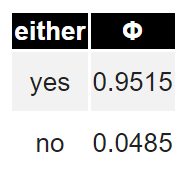}
    \includegraphics[width=0.2\linewidth]{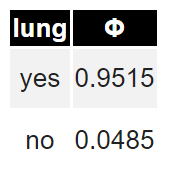}
    \caption{Effects of the argument shown in figure \ref{fig:argument_example}. }
    \label{fig:example_argument_effects}
\end{figure}

\subsection{Argument Strength}

The definition of argument effects is a complete description of how the argument affects our beliefs of all intermediate nodes. But if we are to compare arguments to each other we need a comparable summary of the effects.

For this purposes we define the overall \textbf{strength of an argument} with respect to a target outcome $T = t_o$ as the logarithmic odds in favor of that outcome implied by the effect of the argument on the variable $T$.

\begin{equation*}
\begin{split}
\text{ArgumentStrength}(\mathcal{A}, T=t_o) := \\
\log \frac{\text{ArgumentEffect}(\mathcal{A}, T)(t_o)}{\text{Average}_{t \not =  t_o}\text{ArgumentEffect}(\mathcal{A}, T)(t)}
\end{split}
\end{equation*}

This measure of strength is a real number, whose sign represents whether it is an argument in favor or against the outcome, and whose magnitude represents the amount of evidence provided by the argument.

For example, based on the argument effects in figure \ref{fig:example_argument_effects} we conclude that the strength of the argument for the conclusion $\text{Lung} = \text{yes}$ is $\log\frac{0.9515}{0.0485} = 2.98$.

The definition of $\text{ArgumentStrength}$ corresponds to the strength of the evidence that would be implied by the update associated with the argument if we started from indifference. 

For example, let's suppose we have a variable with three possible outcomes $A,B,C$, and an argument whose effect on the variable is $e_1 : e_2 : e_3$.

If we were initially indifferent between the outcomes, the implied Bayesian update would be:

$$
\underbrace{
\begin{pmatrix}
1 \\ 1 \\ 1
\end{pmatrix}
}_{\text{Prior}}
\times
\underbrace{
\begin{pmatrix}
e_1 \\ e_2 \\ e_3
\end{pmatrix}
}_{\text{Update}}
=
\underbrace{
\begin{pmatrix}
e_1 \\ e_2 \\ e_3
\end{pmatrix}
}_{\text{Posterior}}
$$

The prior odds for outcome $A$ were $1:2$, and the posterior odds are $e_1 : (e_2 + e_3)$. So it makes sense that the strength of the argument should be $\frac{e_1 : (e_2 + e_3)}{1:2} = \frac{e_1}{\frac{e_2 + e_3}{2}}$. We take the logarithm of this quantity to make the multiplicative update an additive one.

\section{Breaking down complex arguments}

Complex arguments can often be broken up into simpler arguments.

\begin{figure}[H]
    \centering
    \includegraphics[width=\linewidth]{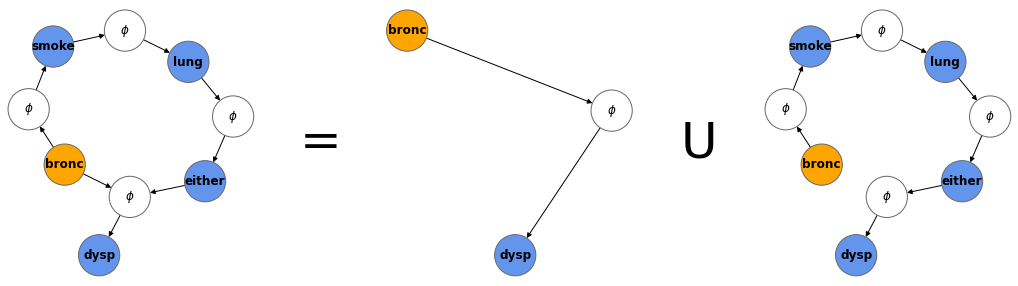}
    \caption{Argument $\mathcal{A}$ from figure \ref{fig:loopy_argument} can be decomposed into subarguments $\mathcal{A}_1$ and $\mathcal{A}_2$.}
    \label{fig:argument_union}
\end{figure}

For example, the loopy argument $\mathcal{A}$ from before can be broken into two subarguments $\mathcal{A}_1$ and $\mathcal{A}_2$, see figure \ref{fig:argument_union}.

In order to produce a good explanation of the effect of the observation `bronc` on the target `dysp` we need to answer one key question: should we present to the user the complex argument $\mathcal{A}$ or the two subarguments $\mathcal{A}_1$ and $\mathcal{A}_2$?

This problem was previously discussed eg in \cite{keppens_argument_2012}, where the author explains the distinction between \textit{convergent arguments} - where each of them independently supports a conclusion - and \textit{linked arguments} - where the strength of each argument depends on the presence of the other.

We need a formal way to decide whether the arguments are independent. If they are, then we can break down the composite argument into its simpler subcomponents. But if the interaction between the arguments is important to support the conclusion we need to present the more complex argument to the user.

First, some preliminaries.

The \textbf{argument union} is defined in the obvious way, as the union of the graphs. it is not well defined when the result would contain a directed loop. We also require that the premises of the joined arguments must be disjoint from the non-premise nodes on other arguments.

We will formally say that an argument $\mathcal{A}$ is a \textbf{subargument} of an argument $\mathcal{A}'$ if every edge, every node and every premise in $\mathcal{A}$ is also in $\mathcal{A}'$. We denote this relation as $\mathcal{A} \subset \mathcal{A}'$.

Lastly, we say that an argument is \textbf{simple} if it cannot be expressed as the union of (distinct) subarguments. An argument is simple iff it consists of a simple path from a single premise to its conclusion. For example, arguments $\mathcal{A}_1$ and $\mathcal{A}_2$ in figure \ref{fig:argument_union} are simple, while $\mathcal{A}$ is not.

Every argument can be expressed as an unique union of simple arguments. We will use this fact later to characterize the set of all arguments. 

\subsection{Argument independence}
Having defined the union of arguments, we are now ready to formally define argument independence.

We say that two arguments are \textbf{independent} if the product of the effects of the arguments is equal to the effect of the union of arguments in said subset.

More generally, we will say that a family of arguments is \textbf{independent} if the product of the effects of any subset of the arguments is equal to the effect of the union of arguments in said subset.

For example, the two arguments $\mathcal{A}_1$ and $\mathcal{A}_2$ we were considering through this section are not independent.

This definition of independence clarifies what it means for two arguments to be independent from the point of view of Bayesian Networks. However, from an explanation point of view the definition is too restrictive.

Because of that we will define a weaker notion of independence, that will help us craft more apt and modular explanations of the network.

We will say that a family of arguments is \textbf{approximately independent} if the product of the effects of any subset of the arguments is within a certain threshold of distance from the effect of the union of arguments in said subset. 

There are many different choices for the factor distance we might use in this definition. Here we opt for one based on the difference of strengths. In this definition, the distance corresponds to the maximum absolute difference in implied logodds between the factors.

\begin{equation*}
\begin{split}
\text{Factor distance}(\phi_1, \phi_2) = \\
\max_{n,s} | \log \frac{(\phi_1 / \phi_2)(t_o)}{\text{Average}_{t \not = t_o}(\phi_1 / \phi_2)(t)} |
\end{split}
\end{equation*}

So for example, while we had that the subarguments $\mathcal{A}_1$ and $\mathcal{A}_2$ were not independent, we find them to be approximately independent within a distance threshold of $0.03$.  

This notion of approximate independence will help us decide if an argument can be broken down into simpler components.

We will say that an argument is \textbf{(approximately) proper} if either it is a simple argument or if it cannot be expressed as the union of a (approximately) independent set of subarguments.

For example, argument $\mathcal{A}$ is $0.03$-approximately proper.

\section{Finding all relevant arguments given some evidence}

We have talked about arguments in a Bayesian Network, and how to identify when it is appropriate to combine arguments or present them separately. What we would like now is a way of, given some input evidence and a target, identify a family of \textbf{relevant and independent arguments} to explain the outcome of the network.

First we will focus on finding all the simple arguments from each evidence node to the target node. This can be achieved with a simple path finding algorithm.

Of the paths, we can discard those which travel along an observed node. Since we already know the value of that node, the arguments up to that point are irrelevant.

\begin{lstlisting}[language=Python, caption=Algorithm to find all simple arguments.]
def all_simple_arguments(model, target, evidence):
  simple_arguments = []
  for path in all_paths(model, evidence, target): 
      if any intermediate path nodes in evidence:   
        continue
      simple_argument = create_argument_from_path(path)
      simple_arguments.append(simple_argument)
  return simple_arguments
\end{lstlisting}


We are finally ready to define and algorithm for finding a list of relevant and independent arguments given some evidence and a target.

Remember that every complex argument can be expressed as the union of simple arguments.

We can use this fact to generate a list of all possible arguments.
For each of these arguments, we can check if the argument can be broken down into an independent combination of arguments.

Of all the proper arguments we are only interested in those which are maximal according to the subargument relationship - we return a list of these, ordered by decreasing absolute strength.

\begin{lstlisting}[language=Python, caption=Algorithm to find all relevant arguments.]
def all_local_arguments(model, target, evidence, threshold=t):
  simple_arguments = all_simple_arguments(model, target, evidence)
  proper_arguments = simple_arguments.copy()
  for components in subsets(simple_arguments):
    argument = compose_argument(components)
    if argument has loops: continue
    for partition in partitions(components):
      subarguments = [compose_argument(subcomponents) for subcomponents in partitions]
      if argument is not t-approximately equal to the combination of the subarguments: 
        proper_arguments.append(argument)
  
  independent_arguments = remove_subarguments(proper_arguments)

  independent_arguments.sort_by_strength()

  return independent_arguments
\end{lstlisting}

\begin{figure}[H]
    \centering
    \includegraphics[width=0.4\linewidth]{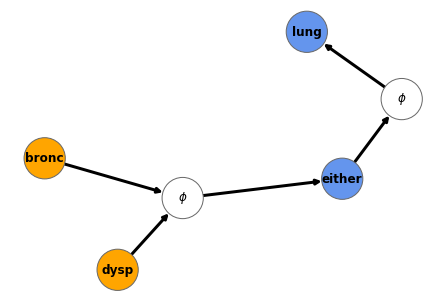}
    \includegraphics[width=0.4\linewidth]{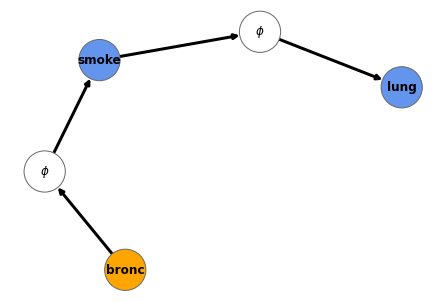}
    \caption{Arguments identified by the argument finding algorithm in the ASIA network when we enter bronc = yes, dysp = no as evidence and lung = yes as target. The dependence threshold is 0.5.}
    \label{fig:example_all_local_arguments}
\end{figure}

Since the number of simple paths in a graph is factorial on the number of variables and the number of complex arguments is exponential on the number of simple paths, the naive approach of listing all arguments and computing all effects will not work in big networks.

Some heuristics can relieve these issues. For example, we can consider only simple paths whose length is below a threshold. Or we can only consider complex arguments which are a combination of up to K simple arguments.

One issue we encountered with these restrictions is that the complexity limits lead us to have too many dependent arguments in the final list, that would have been combined had we allowed for a higher argument complexity. To address this, we refine the final output by iteratively checking for pairwise argument independence in the final outcome, and combine the results.

\begin{lstlisting}[language=Python, caption=Heuristic algorithm to find all relevant arguments.]
def all_local_arguments(model, target, evidence, threshold = t, complexity_limit, max_path_length):
  simple_arguments = all_simple_arguments(model, target, evidence, max_path_length)
  proper_arguments = simple_arguments.copy()
  for components in subsets(simple_arguments, max_subset_size = complexity_limit):
    argument = compose_argument(components)
    for partition in partitions(components):
      subarguments = [compose_argument(subcomponents) for subcomponents in partitions]
      if argument is not t-approximately equal to the combination of the subarguments: 
        proper_arguments.append(argument)
  
  candidate_arguments = remove_subarguments(proper_arguments)

  # Combine dependent arguments
  finished = False
  while not finished:
    finished = True
    for arg1, arg2 in candidate_arguments
      arg = compose_argument(arg1, arg2):
      if arg is not t-approximately equal to the combination of arg1, arg2:
        candidate_arguments.remove(arg1)
        candidate_arguments.remove(arg2)
        candidate_arguments.add(arg)
        finished = False
  independent_arguments = candidate_arguments

  independent_arguments.sort_by_strength()

  return independent_arguments 
\end{lstlisting}

With these restrictions the method can handle medium-size networks, such as ALARM \cite{beinlich_alarm_1989}.

\begin{figure}[H]
    \centering
    \includegraphics[width=\linewidth]{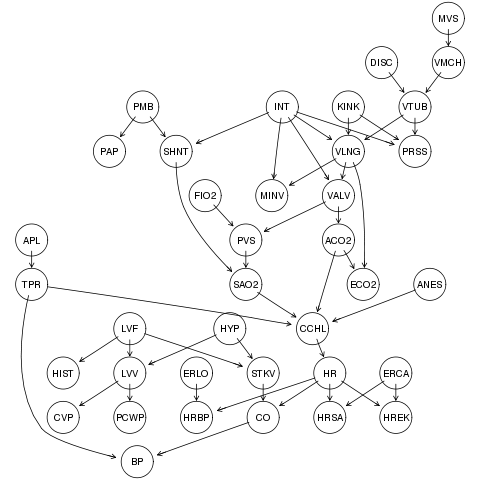}
    \caption{ALARM Bayesian Network. Includes 37 nodes and 509 parameters.}
    \label{fig:alarm_network}
\end{figure}

\begin{figure}[H]
    \centering
    \includegraphics[width = 0.4\linewidth]{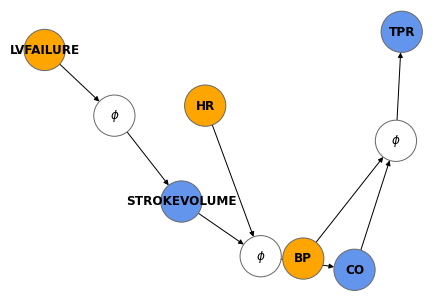}
    \includegraphics[width = 0.4\linewidth]{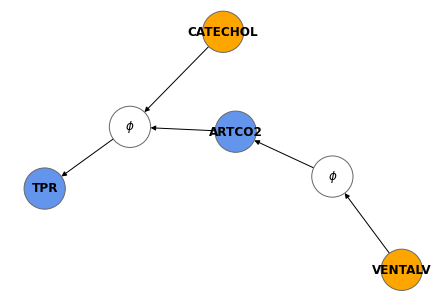}
    \caption{Top 2 arguments identified by the argument finding algorithm in the ALARM network when we enter {'HRSAT': 'HIGH', 'MINVOLSET': 'NORMAL', 'PRESS': 'NORMAL', 'STROKEVOLUME': 'HIGH', 'LVEDVOLUME': 'HIGH', 'HISTORY': 'TRUE', 'HREKG': 'LOW', 'HR': 'NORMAL', 'ERRLOWOUTPUT': 'TRUE', 'CO': 'NORMAL', 'PAP': 'LOW', 'CATECHOL': 'HIGH', 'VENTTUBE': 'ZERO', 'ANAPHYLAXIS': 'FALSE', 'PCWP': 'LOW', 'HRBP': 'HIGH', 'KINKEDTUBE': 'FALSE', 'HYPOVOLEMIA': 'FALSE', 'INTUBATION': 'ONESIDED'} as evidence and 'PULMEMBOLUS', 'FALSE' as target. The dependence threshold is 0.5. As heuristic restrictions we enforce a maximum path length limit of 7 and a maximum combination complexity of 3.}
    \label{fig:alarm_arguments_example}
\end{figure}


We now have shown how to extract a list of relevant and independent arguments from a Bayesian Network, given some evidence and a target outcome.

It is not yet clear whether the arguments are an accurate representation of the reasoning of the Bayesian Network. We address this question in the next section.

\section{Approximating message propagation using arguments and their effects}

Our definition of argument effects is meant to approximate the message passing inference algorithm. We can empirically check whether this is a good approximation.

Concretely, let $\mathcal{A}_1, ..., \mathcal{A}_N$ be the set of relevant, independent arguments found by the argument extraction algorithm in the previous section when presented with evidence $E=e$ and target $T=t_o$.

Then we define the approximate posterior odds of the target node given the arguments as $\hat O(T=t_o | \mathcal{A}_1, ..., \mathcal{A}_N) = O(T=t_o) \cdot \prod_i \text{ArgumentEffect}(\mathcal{A}_i, T)$, where $O(T=t_o)$ are the prior odds of the target as computed by message passing. 

That is, we approximate the posterior odds as the product of the baseline odds and the effects of each relevant argument on the target.

To study the approximation more exhaustively, we will randomly choose evidence and target sets $E=e, T=t_o$ and compare this approximation to the output of message passing. We perform the comparison both in terms of logodds and in term of probabilities.

We also show the error that would be incurred if we approximated every posterior odd as the baseline odds.

\begin{figure*}
    \centering
    \includegraphics[width=\textwidth]{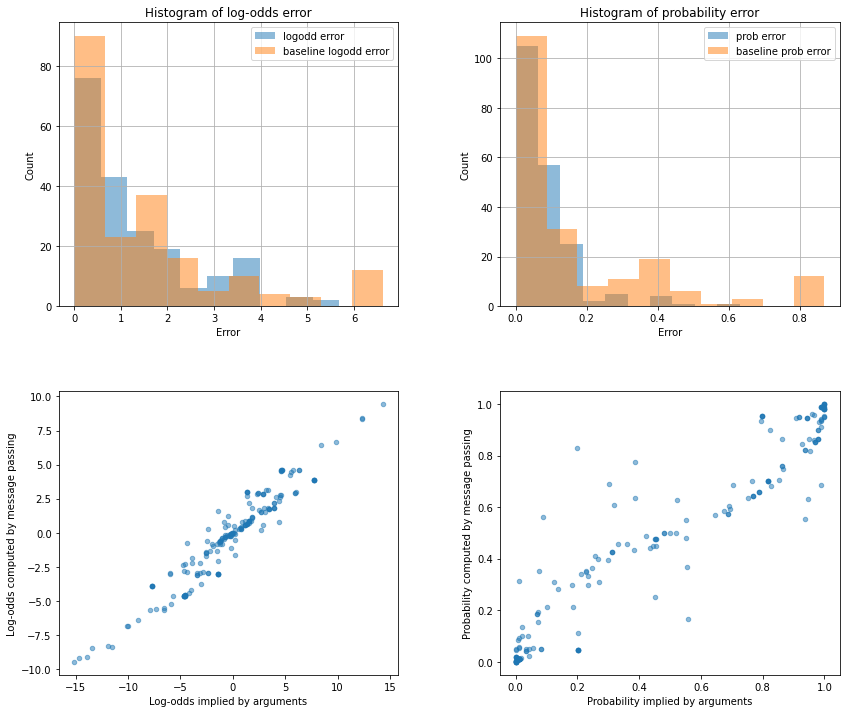}
    \caption{Comparison of message passing and the approximate inference using arguments in the ASIA network. We took n=200 random sets of evidence and targets. In the top row, histogram of absolute errors for logodds (top-left) and for probabilities (top-right). In the bottom row, scatter plots comparing the results of both algorithms.}
    \label{fig:asia_results}
\end{figure*}

Figure \ref{fig:asia_results} shows the summary of this comparison in the ASIA network.

We can see that the results are significantly above the baseline - the arguments are overall shifting the posterior odds in the correct direction. It is not a perfect approximation by any means - but it is qualitatively right.

A visual inspection of the scatter plot showing the logodds implied by the arguments versus the logodds computed by message passing in the ASIA network demonstrate a high correlation.

\begin{figure*}
    \centering
    \includegraphics[width=\textwidth]{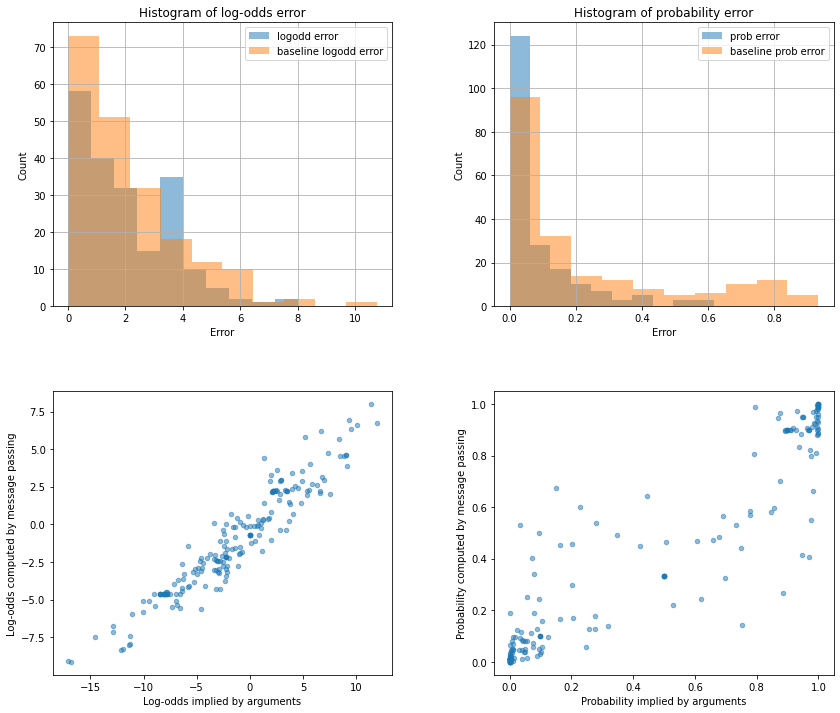}
    \caption{Comparison of message passing and the approximate inference using arguments in the ALARM network. We took n=200 random sets of evidence and targets. In the top row, histogram of absolute errors for logodds (top-left) and for probabilities (top-right). In the bottom row, scatter plots comparing the results of both algorithms.}
    \label{fig:alarm_results}
\end{figure*}

It is also important to explore the results in more complex networks. Figure \ref{fig:alarm_results} shows the summary of this comparison in the ALARM network. In those we use heuristic restrictions described in the previous section to make the problem tractable.

The results are less clear cut in the ALARM network. This could be because of the heuristic restrictions we needed to impose to make the problem tractable.

A summary of the distribution of errors for both the ASIA and the ALARM network can be found in table \ref{table:results_comparison}.

\begin{table*}[]
\begin{tabular}{ccccccccccc}
\hline
\multicolumn{1}{|c|}{\multirow{2}{*}{Network}} & \multicolumn{1}{c|}{\multirow{2}{*}{Method}} & \multicolumn{4}{c|}{Absolute logodd error}                                                                                        & \multicolumn{4}{c|}{Absolute probability error}                                                                                   & \multicolumn{1}{l|}{\multirow{2}{*}{Exact result}} \\ \cline{3-10}
\multicolumn{1}{|c|}{}                         & \multicolumn{1}{c|}{}                        & \multicolumn{1}{c|}{Q 0.05} & \multicolumn{1}{c|}{Median} & \multicolumn{1}{c|}{Q 0.95} & \multicolumn{1}{c|}{Mean} & \multicolumn{1}{c|}{Q 0.05} & \multicolumn{1}{c|}{Median} & \multicolumn{1}{c|}{Q 0.95} & \multicolumn{1}{c|}{Mean} & \multicolumn{1}{l|}{}                              \\ \hline
\multirow{2}{*}{ASIA}                          & Argument                                     & 0.00                               & \textbf{0.77}               & \textbf{3.83}                      & \textbf{1.28}             & 0.00\%                             & \textbf{4.97\%}             & \textbf{27.80\%}                   & \textbf{7.78\%}           & \textbf{8.00\%}                                    \\
                                               & Baseline                                     & 0.00                               & 0.86                        & 5.98                               & 1.49                      & 0.00\%                             & 5.47\%                      & 86.97\%                            & 17.20\%                   & 11.50\%                                            \\
\multirow{2}{*}{ALARM}                         & Argument                                     & 0.02                               & \textbf{1.63}               & \textbf{4.49}                      & \textbf{2.01}             & 0.01\%                             & \textbf{3.43\%}             & \textbf{37.26\%}                   & \textbf{8.55\%}           & 4.00\%                                             \\
                                               & Baseline                                     & \textbf{0.00}                      & 1.62                        & 5.91                               & 2.06                      & \textbf{0.00\%}                    & 10.72\%                     & 81.75\%                            & 22.41\%                   & \textbf{7.00\%}                                   
\end{tabular}
\caption{Summary of the distribution of absolute errors with respect to message passing for random conditional queries in the ASIA and ALARM networks. Argument is our proposed approximation method, baseline is the result when we guess $P(Y=y | X = x) = P(Y=y)$. Each distribution was sampled $n=200$ times. In the exact result column we show the proportion of random queries that perfectly match message passing.}
\label{table:results_comparison}
\end{table*}

\section{Explaining arguments}

So far we have focused on how to select the arguments we want to present to the user.  

In this section we will grapple with the question of how to explain each argument in natural language.

First we will show how to explain a single inference step, and we will build on that to show how to explain a whole argument.

The description of the step depends on:
\begin{itemize}
    \item The type of inference.
    \item The valuation which maximizes the effects of the inference step.
    \item The magnitude of the implied logodds for said valuation.
\end{itemize}

There are three types of inference we need to take into account. 

\begin{itemize}
    \item In a \textbf{causal} inference, the premises are the causal parents of the conclusion. For example, deducing that since the light switch is turned the light must be on.
    \item In an \textbf{evidential} inference, we relate the state of a children to the state of a parent. For example, inferring that there must be fire from seeing smoke come out of the door.
    \item Finally, in an \textbf{intercausal} inference we modulate the effect of an evidential inference by taking into account the presence of another possible cause. For example, usually learning that the floor is wet would lead us to believe that it might have rained. But if we learn that the floor is wet for other reason, eg because of a sprinkler, this would inhibit this line of reasoning.
\end{itemize}

We can deduce the type of inference just from looking at the graphical relation of the premises and the conclusion variables in the model, and explain accordingly.

For a causal or evidential inference, we will use a template like \textit{\{description of the premises\} \{verb\} \{description of the conclusion\} (\{strength qualifier\} inference).}, where the verb is either \textit{causes} or \textit{is evidence that}.

The description of the premises and the conclusion is generated by finding the outcome favored by the evidence, and the description dictionary we defined earlier.

The \textit{strength qualifier} reflects the strength of the inference, and is computed right now in function of the implied logodds of the effect of the rule on the conclusion given the premises. This is similar to the definition of $\text{ArgumentStrength}$.


For an intercausal inference, we offer two options. In `contrastive` mode the description is the same as in the purely evidential case. In `direct` mode, we will compute the counterfactual effect we would have gotten had we observed the premise corresponding to the child of the conclusion node, but not the parents, and we compare that to the actual inference.

We then offer a two part explanation, where we first explain the counterfactual inference, and contrast it with the actual inference.


A natural language explanation of an argument will involve:
\begin{itemize}
    \item Explaining the \textbf{observations} the argument is based on.
    \item Explaining each of the \textbf{inference rules} being applied.
    \item Explaining the \textbf{conclusion} of the argument.
\end{itemize}

For this we will iterate over the argument from observations to its conclusion. The observations are plainly stated, while the rules are explained following the procedure delineated above.

We need to handle the situation where multiple rules affect the same node. In this present article we opt for the option of just chaining together the explanation of each rule, and then adding an extra line explaining the cumulative effect of all the rules.

To the explanation modes we add a third option, \textit{overview}, which will produce a simple description of the observations and the conclusion of the argument.

Let's see some examples. Take for example the argument shown in figure \ref{fig:argument_example}.

Here is a \textbf{direct explanation} of the argument:

\noindent\fbox{\parbox{\linewidth}{
We have observed that the patient does not have tuberculosis and the lung xray shows an abnormality.  \\
That the lung xray shows an abnormality is evidence that the patient has a lung disease (strong inference).  \\
That the patient does not have tuberculosis and the patient has a lung disease is evidence that the patient has lung cancer (strong inference). 
}}
\hfill \break
Here is a \textbf{contrastive explanation} of the argument:

\noindent\fbox{\parbox{\linewidth}{
We have observed that the patient does not have tuberculosis and the lung xray shows an abnormality.  \\
That the lung xray shows an abnormality is evidence that the patient has a lung disease (strong inference). \\  
Usually, if the patient has a lung disease then the patient has lung cancer.    \\
Since the patient does not have tuberculosis, we can be more certain than this is the case (strong inference).
}}
\hfill \break

And here is an \textbf{overview} of the argument:
\noindent\fbox{\parbox{\linewidth}{
Since the patient does not have tuberculosis and the lung xray shows an abnormality, we infer that the patient has lung cancer (strong inference).  
}}
\hfill \break

It will also be illustrative to show the description of an argument in a more complex network with non-binary nodes.

\begin{figure}
    \centering
    \includegraphics[width = \linewidth]{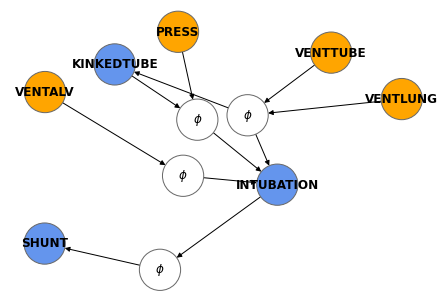}
    \caption{Example of a complex argument in the ALARM network. The premises are that that VENTTUBE is HIGH, VENTLUNG is NORMAL, PRESS is ZERO and VENTALV is HIGH. The target is SHUNT.}
    \label{fig:alarm_example_argument}
\end{figure}

Consider the argument in figure \ref{fig:alarm_example_argument}, corresponding to the ALARM network (see the ALARM network in figure \ref{fig:alarm_network}).

A generated direct description of the argument is:

\noindent\fbox{\parbox{\linewidth}{
We have observed that PRESS is ZERO, VENTLUNG is NORMAL, VENTTUBE is HIGH and VENTALV is HIGH.  \\
That VENTLUNG is NORMAL and VENTTUBE is HIGH is evidence that KINKEDTUBE is FALSE (strong inference).  \\
That PRESS is ZERO and KINKEDTUBE is FALSE is evidence that INTUBATION is ESOPHAGEAL (strong inference).  \\
That VENTLUNG is NORMAL and VENTTUBE is HIGH is evidence that INTUBATION is ESOPHAGEAL (strong inference).  \\
That VENTALV is HIGH is evidence that INTUBATION is NORMAL or INTUBATION is ESOPHAGEAL (strong inference).  \\
All in all, this is evidence that INTUBATION is ESOPHAGEAL (strong inference).  \\
That INTUBATION is ESOPHAGEAL causes that SHUNT is NORMAL (moderate inference).  
}}

\section{Conclusion}

In this article we have introduced several building blocks - arguments over the factor graph; the notion of link effects, argument effects and argument strength; the notion of subarguments, argument union and argument (approximate) independence; and the notions of simple and proper arguments.

We used all these concepts to create an algorithm which given a Bayesian Network, a target outcome and some evidence will output a list of relevant and independent arguments summarizing the different ways the evidence affects the output.

We have shown that we can get an approximation of the posterior odds based on the strength of the extracted arguments, which hints at the accuracy of the approach.

And we have further shown a simple scheme to explain the arguments using natural language.

My work here has set up a foundation for a mathematical theory of explanations in Probabilistic Graphical Models. However, it is not clear this is the best approach to the topic.

Possible next steps include:

\begin{itemize}
    \item Applying the methods in a user study, comparing it to previous approaches for explaining Bayesian Networks.
    \item Finding more efficient implementations of the algorithms in the paper, that could handle networks of large size ($>100$ nodes).
    \item Improving the methods for generating the textual and visual explanations of the arguments.
    \item Comparing the variable selection implicitly realized by the argument generation to other variable selecting schemes.
\end{itemize}

\section{Acknowledgements}

This work has received funding from the European
Union’s Horizon 2020 research and innovation programme under the Marie Skłodowska-Curie grant
agreement No 860621.

I thank Tjitze Rienstra, Ehud Reiter, Oren Nir, Nava Tintarev, Pablo Villalobos, Michele Cafagna, Ettore Mariotti, Sumit Srivastava, Alberto Bugarín and Ingrid Zuckerman for feedback, discussion and support. 

\appendix


 \bibliographystyle{elsarticle-num} 
 \bibliography{bayesian_networks}





\end{document}